\ifdefined\XeTeXversion
  \RequirePackage[OT1]{fontenc}
\fi
\documentclass[letterpaper,10pt,conference]{ieeeconf}
\IEEEoverridecommandlockouts
\usepackage{times,graphicx,amsmath,amssymb,booktabs,array,cite,etoolbox}
\usepackage{stfloats,capt-of,float}
\usepackage[hidelinks]{hyperref}

\newcommand{\wrap}{\operatorname{wrap}}
\newcommand{\tablecaptiontext}[1]{\begingroup\leftskip=0pt\rightskip=0pt\parfillskip=0pt\tolerance=10000 #1\par\endgroup}
\makeatletter
\let\FortuneOriginalMakeCaption\@makecaption
\long\def\@makecaption#1#2{%
 \ifx\@captype\@IEEEtablestring
  \FortuneOriginalMakeCaption{#1}{#2}%
 \else
  \@IEEEfigurecaptionsepspace
  \parbox[t]{\hsize}{\footnotesize\raggedright\parindent=0pt\noindent #1.\enspace #2\par}%
 \fi}
\makeatother
\title{\normalfont\fontsize{16.5pt}{20pt}\selectfont\textbf{\FortuneProject}: \FortuneSubtitle}
\def\FortuneAuthorNames{Philip Niu, Xiaofeng Han, Sylvie Pan, Yang Wang}
\hypersetup{pdfauthor={\FortuneAuthorNames}}
\newcommand{\FortuneFooterAffiliation}{\textsuperscript{1}\,Fortune AI}
\newcommand{\FortuneFooterContact}{Corresponding author: Yang Wang, \href{mailto:wangglenwang@gmail.com}{wangglenwang@gmail.com}.}
\author{Philip Niu\textsuperscript{1}, Xiaofeng Han\textsuperscript{1}, Sylvie Pan\textsuperscript{1}, Yang Wang\textsuperscript{1\,\href{mailto:wangglenwang@gmail.com}{\includegraphics[height=.68em]{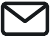}}}\hspace{.21em}\href{https://orcid.org/0009-0004-0130-8338}{\raisebox{.12ex}{\includegraphics[height=.9em]{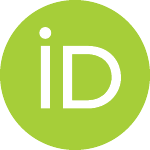}}}\ifdefined\FortuneSubmission\thanks{\FortuneFooterAffiliation\newline\FortuneFooterContact}\fi}
\def\FortuneProject{MotionSkill}
\def\FortuneSubtitle{From Bipedal Motion Video to Control across Embodiments}
\ifdefined\FortuneSubmission\else
 \newsavebox{\FortuneLogoBox}
 \sbox{\FortuneLogoBox}{\includegraphics[width=21pt,height=21pt,keepaspectratio]{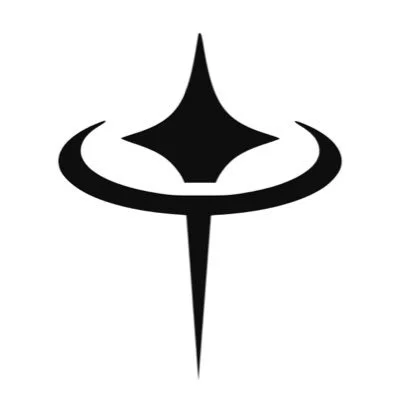}}
 \newcommand{\FortuneHeaderLogo}{\usebox{\FortuneLogoBox}}
 \input{fortune-paper.sty}
\fi
\ifdefined\FortuneSubmission\else
\apptocmd{\thebibliography}{\setlength{\itemsep}{1.8pt}}{}{}
\fi
\renewcommand{\footnoterule}{\kern-3pt\hrule width .35\columnwidth height .4pt\kern 2.6pt}
\begin{document}
\maketitle
\thispagestyle{plain}\pagestyle{plain}
\begin{abstract}
Reusing a demonstrated gait can reduce the work of specifying the same motion for each biped. A video captures the timing of a step, but each robot must execute it with its own limbs and actuators. MotionSkill recovers a timed reference once from monocular video and adapts it to each target in simulation. It aligns root motion and body landmarks in a common frame and records which landmarks are available at each instant. At each source timestamp, the mapper preserves limb directions, uses the target segment lengths, and solves inverse kinematics under joint limits and closure constraints. In MuJoCo, we train a separate residual policy for each of ten bipeds on its mapped reference from one four-second walking clip. Each policy uses the measured robot state to correct joint commands during contact. From initial states held out during training, MotionSkill completed 98\% of trials on the original video and 26\% with the upper body occluded for one second.
\end{abstract}

\section{Introduction}
A video shows a person shifting weight and advancing one foot. A compact biped and a robot with longer legs or a closed knee linkage need different joint trajectories for that step. Both can use the directions of the limbs and the order and timing of foot motion.

In MotionSkill, a skill is the timed motion reference recovered once from a demonstration. To deploy it on another body, we reuse the same samples and timestamps, map them to that robot, and train its feedback policy. Limb directions specify what to match. Segment lengths and joint constraints determine the target pose at each instant. Physical contact and feedback determine when the robot actually reaches it.

We use one filmed walk across ten robots. Nine complete all five trials on the original input, and the tenth completes four. With the upper body masked for part of the clip, we measure how many robots receive a feasible joint trajectory and how many trials reach the end of that trajectory. A longer recorded walk tests repeated stepping, and a change in source cadence tests how closely the feet follow the demonstrated rhythm.

The work makes three contributions.
\begin{itemize}
\item A timed representation of root motion, landmark geometry, and point availability that can be recovered once per video and reused across target bodies.
\item A mapping and control procedure that preserves native segment lengths and closure constraints and uses reference conditioned joint residuals to track the motion in simulation.
\item Paired video trials that measure mapping feasibility and execution on ten bipeds, and a cadence experiment that measures foot timing on four robots using their pretrained locomotion controllers.
\end{itemize}

\begin{figure*}[!t]
\centering\includegraphics[width=\textwidth]{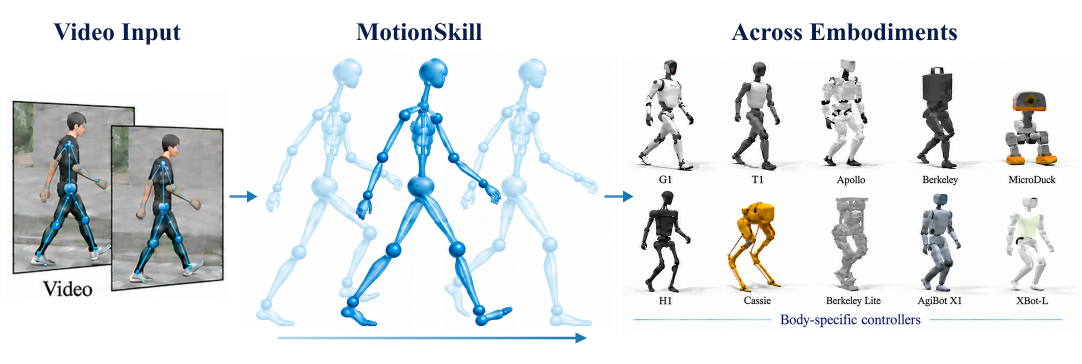}
\caption{\textbf{MotionSkill overview.} A walking video supplies one timed motion reference. The three poses show the direction of travel. Each of the ten simulated bipeds maps that reference to its own joint geometry and tracks it with a separately trained controller.}
\label{fig:overview}
\end{figure*}

\section{Related Work}
\subsection{Motion representation and recovery}
SMPL and AMASS provide common representations for human motion~\cite{loper2015smpl,mahmood2019amass}. WHAM, GVHMR, and MoRo estimate motion from video~\cite{shin2024wham,shen2024gvhmr,qian2026moro}. GMR maps a motion sequence to robot configurations, while UniMate transfers animation across joint topologies~\cite{araujo2026gmr,mou2026unimate}. VideoMimic connects video reconstruction to physical control~\cite{allshire2025videomimic}. ZEST trains tracking policies in simulation and transfers skills recovered from video to Atlas and Unitree G1 hardware~\cite{sleiman2026zest}. Motion imitation, learned priors, and motor adaptation provide ways to realize reference motion in simulation~\cite{peng2018deepmimic,peng2021amp,luo2024pulse,luo2023phc,tessler2024maskedmimic,kumar2021rma}. We follow one video estimate through mapping and physical execution on ten bipeds.

\subsection{Cross embodiment mapping and control}
MetaMorph and Body Transformer encode morphology for control~\cite{gupta2022metamorph,sferrazza2024body}. MorphMimic shares a world model and tracking policy conditioned on morphology~\cite{yang2026morphmimic}. MimicAll conditions a shared tracker on body descriptors, X-WBC aligns motion tokens, and Any2Any combines kinematic alignment with dynamic adaptation~\cite{li2026mimicall,zhang2026xwbc,yang2026any2any}. In MotionSkill, the quantity transferred between bodies is the timed task reference. Each body receives the same segment directions and solves for its own joint configuration. We check the resulting trajectory against the target's kinematic constraints before passing it to the controller.

\subsection{Reference conditioned control and timing}
Phase oscillator learning couples robot progression to moving targets~\cite{maeda2018phase}. AdaMimic adapts edited motions~\cite{huang2025adamimic}, and HuGWBC exposes gait frequency and phase as commands~\cite{xue2025hugwbc}. ExBody, HumanPlus, H2O, and OmniH2O use human motion to guide humanoid control~\cite{cheng2024exbody,fu2024humanplus,he2024h2o,he2024omnih2o}. ASAP, BeyondMimic, TWIST2, and SuperMimic develop tracking policies and deployment methods~\cite{he2025asap,liao2025beyondmimic,ze2025twist2,liao2026supermimic}. We study timing at the controller input by fixing the policy and varying when it receives the phase of the demonstrated gait.

\section{The MotionSkill Framework}
\subsection{Framework and motion representation}
MotionSkill represents the demonstration independently of the target joint layout. For video, reference preparation estimates body geometry and global motion on the source clock. A recorded motion can supply these quantities directly. At time $t$, the skill is
\begin{equation}
S(t)=\{p_t,R_t,J_t,v_t,\omega_t,\phi_t,m_t\},
\label{eq:contract}
\end{equation}
Here $p_t,R_t$ specify root motion, $J_t$ contains named body points, and $v_t,\omega_t$ specify travel. Periodic motion can also supply gait phase $\phi_t$. These channels share source time, allowing a controller to relate the desired pose to the requested progression. The availability state $m_t$ identifies each point as observed, recovered, imputed, or invalid. When observations contain a short interior gap, valid endpoints support temporal repair before the geometry enters mapping. The repaired points keep their imputation labels on the same timeline as the observed points.

At each source time $t$, the mapper constructs $\hat Q_r(t)$ from the same $S(t)$ for every target $r$ (Sec.~\ref{sec:mapping}). Geometry and joint commands vary by body at a shared requested time. Feedback uses the measured state $o_t^r$ to track that request (Sec.~\ref{sec:tracking}), so tracking lag and stride differences remain observable.

\subsection{Source alignment}
Source alignment converts the observed sequence into the coordinates and timestamps required by the skill. It preserves the availability of each channel so that mapping can identify the geometry it can use. Video provides estimated points and root motion, while recorded sequences provide the same quantities for controlled component tests.

\textbf{Recorded pose sequences.} For a point $p_{j,t}$ in a source trajectory, root position $p_t$, orientation $R_t$, and initial height $h_0$, we form
\begin{equation}
\bar p_{j,t}=R_t^\top(p_{j,t}-p_t)/h_0.
\label{eq:canonical}
\end{equation}
Root velocity and yaw rate supply travel. Maxima of normalized ankle separation define gait cycles, with phase advancing by $2\pi$ between maxima. Sagittal arm and thigh angles and knee flexion supply geometry. Derivatives and smoothing produce channels at physical time.

\textbf{Video estimates.} Global motion and camera frame pose are estimated from the same source frames~\cite{shen2024gvhmr,qian2026moro}. Let $R_w(t),R_c(t)$ be the global estimator's world and camera root rotations, $z(t)$ its world translation parameter, and $J^c_j(t)$ the camera frame position of joint $j$. With $B$ converting to a right handed frame, the source adapter forms
\begin{equation}\begin{aligned}
Q_t&=R_w(t)R_c(t)^\top,\\
p_t^U&=p_0+B[z(t)-z(0)],\\
J^U_{j,t}&=BQ_t[J^c_j(t)-J^c_{\mathrm{pelvis}}(t)]+p_t^U.
\end{aligned}\label{eq:video}
\end{equation}
The transformation expresses pelvis-relative geometry in the common frame and retains estimated world displacement and orientation. Each sample carries its source timestamp into mapping, including poses recovered when part of the body is occluded.

\begin{figure*}[!t]
\centering\includegraphics[width=\textwidth]{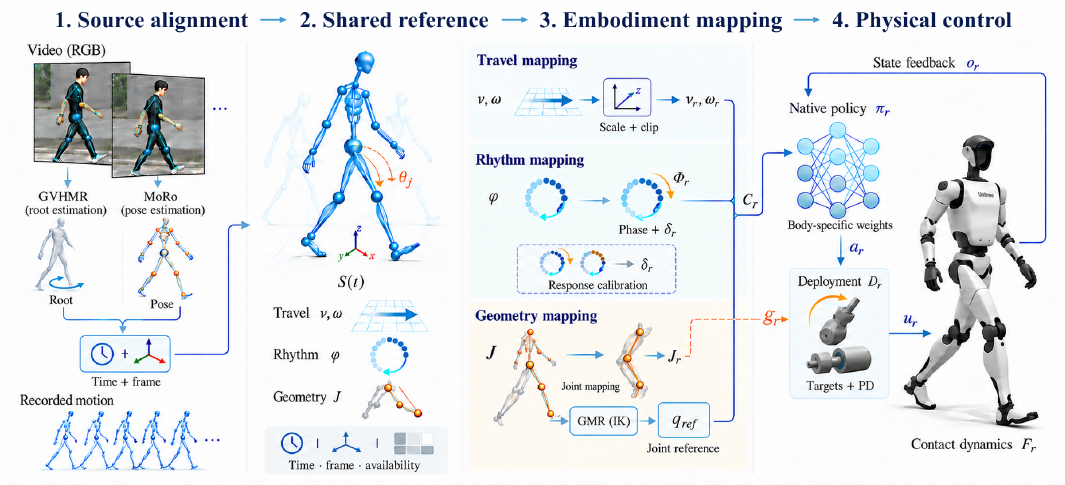}
\caption{\textbf{MotionSkill framework and controller adapters.} Source alignment supplies root motion, body points, and time. Mapping adapts these quantities to the target geometry. Joint references carry recorded and recovered motion into physical tracking. Travel and rhythm commands support the retained controller timing study. State feedback closes the physical loop in each case. Native limits and passive joints remain in the target model.}
\label{fig:framework}
\end{figure*}

\subsection{Embodiment mapping}
\label{sec:mapping}
MotionSkill maps the skill's segment directions onto the target body and retains that body's segment lengths. Let $J_i(t)$ and $J_j(t)$ be two connected skill points. The term $\ell^r_{ij}$ is the native length of the corresponding target segment, and $H_r$ aligns the initial headings of the skill and target. The mapped point satisfies
\begin{equation}
\hat J^r_j(t)=\hat J^r_i(t)+\ell^r_{ij}H_r
\frac{J_j(t)-J_i(t)}{\|J_j(t)-J_i(t)\|}.
\label{eq:segment}
\end{equation}
The pelvis is anchored at the midpoint of the two hip points, and its displacement is scaled by the ratio of target to source leg length. Constrained inverse kinematics converts these targets into native joint configurations~\cite{araujo2026gmr}. These problems retain joint limits, passive joints, and closed chain constraints, and use the preceding solution to keep the mapped motion continuous.

The result is a native reference
\begin{equation}
\hat Q_r(t)=\{\hat q_t^r,\hat p_t^r,\hat R_t^r,\hat v_t^r\},
\label{eq:native}
\end{equation}
with actuated joint positions, rigid body poses, and velocities on the source timeline. Mapping uses valid recovered or repaired points, whose availability states preserve their origin. This reference connects the common skill to the target controller.

\subsection{Reference conditioned physical tracking}
\label{sec:tracking}
A mapped pose specifies a configuration. Balancing at that pose also depends on velocity, ground contact, and the deviation accumulated on preceding steps. For body $r$, the policy $\pi_{\theta_r}$ observes proprioception, the current reference command, the reference anchor pose relative to the measured body, the previous action, and reference velocity in the measured anchor frame. The velocity input is scaled by native leg length.

The policy predicts a residual around the mapped actuated joint reference,
\begin{equation}\begin{aligned}
a_t^r&=\pi_{\theta_r}\!\left(o_t^r,\hat Q_r(t)\right),\\
q^{r,\mathrm{tar}}_t&=\hat q^r_{\mathrm{act},t}+\mathcal B_r(\Lambda_r a_t^r),
\end{aligned}\label{eq:policy}
\end{equation}
where $\hat q^r_{\mathrm{act},t}$ selects the actuated reference joints, $\Lambda_r$ is the native action scale, and $\mathcal B_r$ bounds selected posture residuals. Joint feedback and the native actuator model convert the target into physical control. The floating base, passive joints, and closed chains evolve under contact dynamics. Body geometry, actuator parameters, and policy dimensions follow the target model. The observation definitions, residual action rule, and training objective are shared.

PPO~\cite{schulman2017ppo} optimizes pose, root motion, and velocity tracking with penalties for action variation, joint limits, and self contact. Planar velocity and end effector terms are normalized by native leg length, making travel and distal correspondence comparable across bodies. Angular terms retain common settings.

\subsection{Timing for command conditioned controllers}
For these pretrained controllers, which accept velocity and phase commands, the same reference supplies a timing adapter. The controller receives travel commands from $v_t,\omega_t$ and a phase $\phi_t+\delta_r$. The native clock condition aligns the initial phase once and then uses the controller's own clock. The continuous condition reads source phase before every policy observation. Both conditions use the same root commands and geometric inputs at each requested source time.

The offset $\delta_r$ accounts for differences in how controllers realize a phase command. Two physical probes apply offsets of $0$ and $\pi/2$. For each probe, measured ankle separation is fitted against source phase as
\begin{equation}
\begin{aligned}
d_r(t)&=a\cos\phi_t+b\sin\phi_t+c,\\
\psi&=\operatorname{atan2}(-b,a).
\end{aligned}
\label{eq:phasefit}
\end{equation}
The same fit gives the source response phase $\psi_S$. With $\psi_0$ and $\psi_{\pi/2}$ denoting the two measured response phases, the local response slope and command correction are
\begin{equation}
\begin{aligned}
g_r&=\frac{\wrap(\psi_{\pi/2}-\psi_0)}{\pi/2},\\
\delta_r&=\wrap\!\left(\frac{\wrap(\psi_S-\psi_0)}{g_r}\right).
\end{aligned}
\label{eq:phaseoffset}
\end{equation}
The fit excludes the initial settling interval and requires a stable probe, measurable periodic motion, and a nondegenerate response slope. A third probe validates the correction before held out evaluation. The resulting calibration changes the phase command while preserving policy weights and root commands.

\section{Experiments}
\subsection{Protocol, baselines, and metrics}
The ten-body walking study uses a separate residual tracker per body. The timing study uses four pretrained locomotion controllers with fixed weights. The separate G1 whole-body example uses a pretrained sequence tracker with its native lookahead~\cite{ze2025twist2}. For the ten-body video study, we use a four second segment of a handheld walking video from the WHAM examples~\cite{shin2024wham}. One input retains all frames. The paired input masks the upper body between $t=1$ and $2$~s. Each input produces a reference with 199 samples at 50~Hz, shared across ten targets. The simulations preserve floating base dynamics, gravity, and contact using MuJoCo and MuJoCo Playground~\cite{mujoco,playground}.

\begin{table}[H]
\caption{\tablecaptiontext{Evaluation design. Each video condition uses a single recovered reference and the same ten target bodies. Video results use held out initial states. The longer recorded walk and native policy trials measure stepping and timing.}}
\label{tab:protocol}\centering\small
\begin{tabular}{@{}p{.42\columnwidth}p{.48\columnwidth}@{}}\toprule
Study & Design\\\midrule
Video deployment & 10 bodies, 2 inputs, 5 paired reset seeds per body\\
Recorded motion control & 10 targets $\times$ 10 initial states $\times$ 20~s = 100 trials\\
Supporting analyses & 184 timing trials and one original/occluded video pair\\\bottomrule
\end{tabular}
\end{table}

\textbf{Policy adaptation and evaluation.} We train one policy per body on the reference recovered from the original clip, using 3,000 PPO updates and 1,024 parallel environments per body. We fix the final weights and evaluate five held out reset seeds per body and input. Seeds pair perturbations around each condition's initial reference pose. The direct PD baseline sets the learned residual to zero. It tracks each feasible reference with the same joint targets, gains, robot model, and initial states as the learned controller. A trial counts as complete when it reaches the end of the reference with finite states and actions, finite actuator forces within native limits, and no physical termination. Tracking stops if the absolute height error of the anchor or any monitored end effector exceeds $0.25(L/0.60)$~m, where $L$ is the native leg length in metres, or if the vertical components of gravity projected into the reference and measured anchor frames differ by more than $0.8$ in normalized units. The evaluator logs planar drift separately. Mapping failures count toward the planned total.

\textbf{Timing controls.} For the four native policies, native velocity (NV) supplies travel commands. The native clock (NC) condition adds geometry and aligns phase at reset. Continuous source phase (M) also refreshes phase at every observation. Paired trials use the same weights, geometric inputs, and initial states. The calibration baseline uses the fixed phase offset already stored for each body in the evaluation protocol (the archived offset).

\begin{figure*}[!t]
\centering\includegraphics[width=\textwidth]{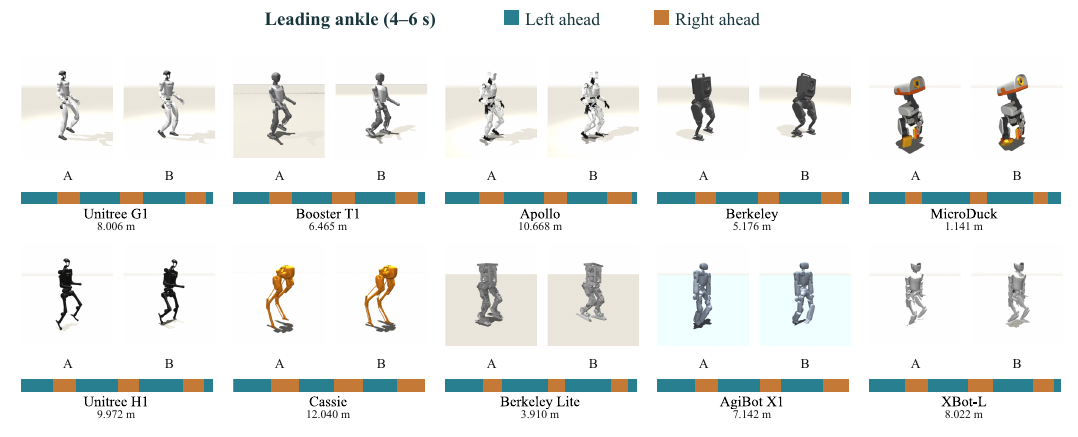}
\caption{\textbf{Stepping poses and leading ankle across ten bodies.} A and B mark two recorded poses of each robot during the 20~s walking trial. The strip below each pair shows which ankle is farther forward from $4$ to $6$~s. Teal marks the left ankle and ochre the right. Distances give mean planar displacement over ten trials per body.}
\label{fig:synchronization}
\par\bigskip\noindent\begin{minipage}[t]{\columnwidth}\normalfont\normalsize\raggedright The 20~s recorded G1 walk tests repeated stepping across the same ten bodies. All 100 trials finish, with 52 to 61 leading foot changes per trial. Mean planar travel ranges from 1.141~m on MicroDuck to 12.040~m on Cassie. The common timeline accommodates these different travel distances.\end{minipage}\hfill\begin{minipage}[t]{\columnwidth}\normalfont\normalsize\raggedright Figure~\ref{fig:synchronization} shows which ankle leads. Figure~\ref{fig:tenbodies} compares the continuous separation with the mapped reference for each robot. A shifted zero crossing indicates a timing difference. The peaks measure how far apart the ankles move along the forward axis. H1 exhibits intermittent lower leg contact during this walk.\end{minipage}\par\bigskip
\centering\includegraphics[width=\textwidth]{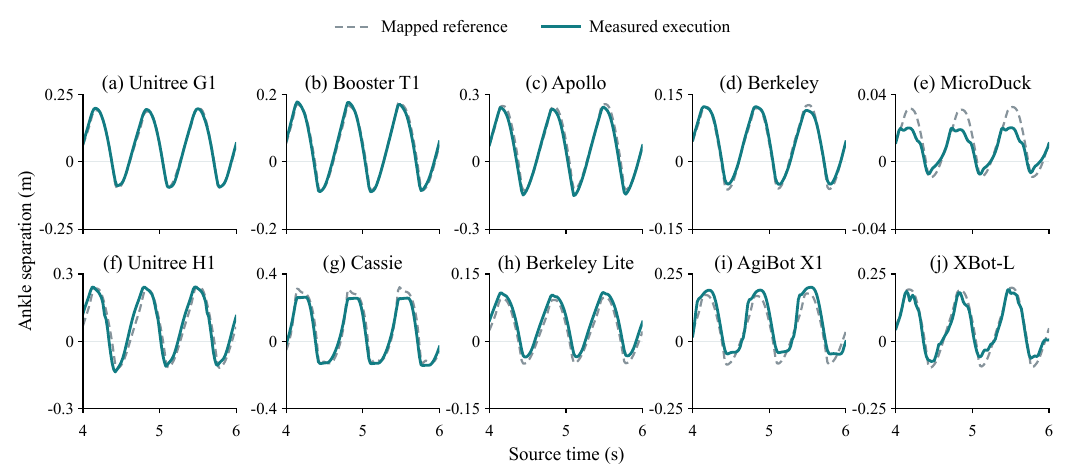}
\caption{\textbf{Executed correspondence across ten bodies.} Solid curves measure left foot position minus right foot position along the root forward axis. Dashed curves show the mapped native reference. Vertical ranges vary by body and include zero. Original timestamps expose correspondence and drift at the requested source times and preserve each robot's spatial scale.}
\label{fig:tenbodies}
\end{figure*}

\textbf{Motion measurements.} For each completed video trial we project anchor travel and ankle separation onto the reference direction of travel. A leading foot change occurs when signed separation crosses to the opposite side of a hysteresis band of $0.03$ leg lengths. After scoring completion, we check for positive travel along the reference direction and at least two leading foot changes. In the timing study we measure planar velocity error and correlation of foot separation at simultaneous samples,
\begin{equation}
\rho_r=\operatorname{corr}(d_r(t),d_S(t)),\qquad t\geq2\ \mathrm{s},
\label{eq:corr}
\end{equation}
where $d_r$ and $d_S$ are the left minus right foot separations along the target and source forward axes. Correlation measures agreement at the requested source times.

\subsection{Physical execution across embodiments}
On the original walking video, MotionSkill completes 49 of 50 trials from held out initial states, compared with 0 of 50 for direct PD (Table~\ref{tab:selected}). The PD trials end within 0.92~s. XBot-L completes four trials. Its remaining trial ends at 1.18~s. Across both video inputs, every completed trial advances along the reference direction and changes the leading foot at least twice. Completed trials on the original video show four to nine changes. A separate pilot applied trackers trained on the recorded walk to the video reference. With one reset for each body, these trackers completed 0 of 10 trials.

\begin{table}[H]
\caption{\tablecaptiontext{Physical execution on ten bipeds. Clean and Mask give completions out of five held out initial states on one video. In cells marked $^{*}$, mapping failed before simulation. All five planned trials count as failures. For the 20~s recorded reference, Alt. gives the range of leading foot changes and Travel gives the mean planar displacement of ten trials per body. All ten trials complete on every target.}}
\label{tab:selected}\centering\small
\setlength{\tabcolsep}{3pt}%
\begin{tabular}{@{}lcccc@{}}\toprule
Body & Clean & Mask & Alt. & Travel (m)\\\midrule
G1 & 5/5 & 0/5 & 59 & 8.006\\
T1 & 5/5 & 0/5$^{*}$ & 58 & 6.465\\
Apollo & 5/5 & 1/5 & 59 & 10.668\\
Berkeley & 5/5 & 0/5 & 58--59 & 5.176\\
MicroDuck & 5/5 & 0/5 & 52--58 & 1.141\\
H1 & 5/5 & 4/5 & 57--59 & 9.972\\
Cassie & 5/5 & 0/5$^{*}$ & 58 & 12.040\\
Berkeley Lite & 5/5 & 5/5 & 59 & 3.910\\
AgiBot X1 & 5/5 & 0/5 & 60--61 & 7.142\\
XBot-L & 4/5 & 3/5 & 59--60 & 8.022\\
\bottomrule
\end{tabular}

\end{table}

With one second of upper-body occlusion, eight of the ten bodies receive feasible references. The root rotation step for T1 and the landmark error for Cassie exceed their respective mapping thresholds, accounting for 10 of the 50 planned trials. On the 40 executable trials, the policies trained on the original input complete 13 (32.5\%) and direct PD completes 0. Across all planned trials, MotionSkill's completion rate is 26\%. The 13 completions comprise five on Berkeley Lite, four on H1, three on XBot-L, and one on Apollo. 

Figures~\ref{fig:synchronization} and~\ref{fig:tenbodies} examine how the same bodies reproduce a separate 20~s recorded walk over multiple strides. G1 and Berkeley Lite each change the leading foot 59 times in every trial. Over the same 20~s, their mean planar displacements are 8.006 and 3.910~m, respectively.

\subsection{Source timing on native controllers}
All 120 fixed reference trials complete. T1 and Berkeley track the aligned native clock closely. Apollo benefits from full phase conditioning, and G1 is slightly better with the aligned native clock at both friction settings. In each changing cadence trial, the source switches from $0.8$ to $1.2$ times its recorded rate at 10~s.

\begin{figure*}[!t]
\centering\includegraphics[width=\textwidth]{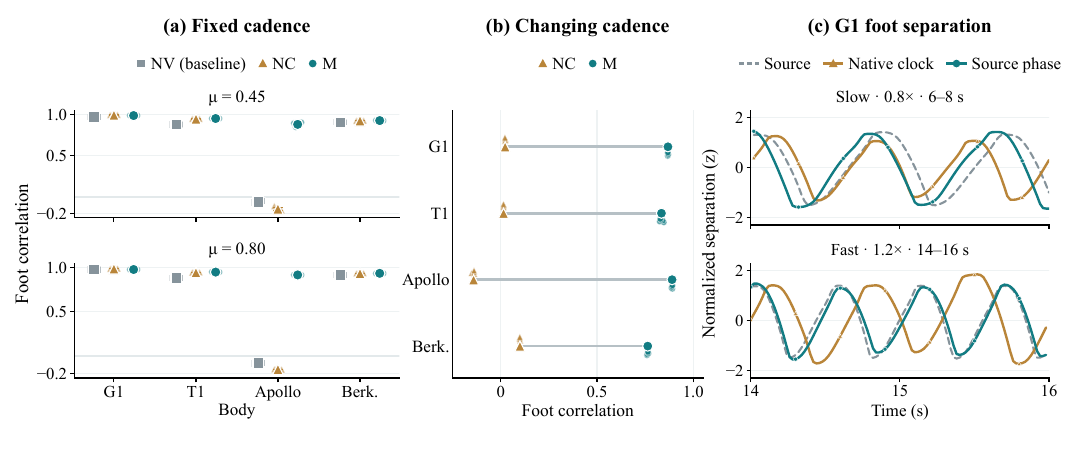}
\caption{\textbf{Timing analysis on retained native controllers.} (a) We compare native velocity (NV), geometry with a native clock (NC), and continuous source phase (M) across 120 fixed cadence trials and two friction settings. (b) Forty changing cadence trials isolate continuous reference timing. Paired comparisons use the same initial state and native controller. Points show all five seeds with mean and sample standard deviation. (c) Standardized foot separation for G1 (seed 0). The dashed gray line is the source reference. Ochre triangles (NC) and teal circles (M) show the measured response under the native clock and continuous source phase, respectively.}
\label{fig:cadence}

\par\medskip\noindent\begin{minipage}[t]{\columnwidth}\normalfont\normalsize\raggedright We set $s(t)=0.8\min(t,10)+1.2\max(t-10,0)$ in both conditions, using geometry at $s(t)$ and root commands scaled by $\dot s(t)$. With continuous phase, all 20 trials succeed. Mean $\rho$ rises from $-0.1411$--$0.0996$ under NC to $0.7635$--$0.8898$, and velocity RMSE falls on all four bodies (Table~\ref{tab:cadence}). Under NC, T1 ends behind its starting position in all five 20~s trials.\end{minipage}\hfill\begin{minipage}[t]{\columnwidth}\normalfont\normalsize\raggedright In the calibration test, both members of each pair receive source phase. Only the offset changes. All 24 trials completed across 12 matched pairs. Measured offsets raised mean foot correlation on all four bodies (Table~\ref{tab:phase}). For T1, the archived and measured offsets give correlations of 0.9161 and 0.9271 and mean forward displacements of 1.323 and 0.775~m, respectively. For Apollo, the corresponding velocity errors are 0.0873 and 0.0955~m/s.\end{minipage}
\par\medskip\noindent\begin{minipage}[t]{\columnwidth}\vspace{0pt}
\captionof{table}{\tablecaptiontext{Changing cadence. All 40 trials ran for 20~s. Foot correlation $\rho$ and planar velocity RMSE use $t\geq2$~s through trial end, including failures. Entries give mean $\pm$ sample SD over five paired seeds. All five T1 NC trials missed the 0.5~m forward criterion.}}
\label{tab:cadence}\centering\small\setlength{\tabcolsep}{2pt}
\begin{tabular}{@{}lcccc@{}}\toprule
Body & Clock & \shortstack{Success\\(\%) $\uparrow$} & $\rho\uparrow$ & \shortstack{Vel. RMSE\\(m/s) $\downarrow$}\\\midrule
G1 & NC & 100 & $0.0229\mathord{\pm}0.0020$ & $0.2078\mathord{\pm}0.0009$\\
 & M & 100 & $0.8684\mathord{\pm}0.0012$ & $0.1524\mathord{\pm}0.0014$\\
T1 & NC & 0 & $0.0144\mathord{\pm}0.0009$ & $0.1807\mathord{\pm}0.0002$\\
 & M & 100 & $0.8348\mathord{\pm}0.0093$ & $0.1147\mathord{\pm}0.0014$\\
Apollo & NC & 100 & $-0.1411\mathord{\pm}0.0059$ & $0.1771\mathord{\pm}0.0007$\\
 & M & 100 & $0.8898\mathord{\pm}0.0022$ & $0.0984\mathord{\pm}0.0005$\\
Berkeley & NC & 100 & $0.0996\mathord{\pm}0.0011$ & $0.2430\mathord{\pm}0.0004$\\
 & M & 100 & $0.7635\mathord{\pm}0.0030$ & $0.1822\mathord{\pm}0.0002$\\
\bottomrule\end{tabular}
\end{minipage}\hfill\begin{minipage}[t]{\columnwidth}\vspace{0pt}
\captionof{table}{\tablecaptiontext{Phase response on held out states with archived (E) or measured (R) offsets. Foot correlation $\rho$, velocity RMSE and forward displacement are reported as mean $\pm$ sample SD.}}
\label{tab:phase}\centering\small\setlength{\tabcolsep}{2pt}
\begin{tabular}{@{}lcccc@{}}\toprule
Body & Offset & $\rho\uparrow$ & \shortstack{Vel. RMSE\\(m/s) $\downarrow$} & \shortstack{Forward\\(m)}\\\midrule
G1 & E & $0.9481\mathord{\pm}0.0008$ & $0.1310\mathord{\pm}0.0009$ & $9.627\mathord{\pm}0.061$\\
 & R & $0.9593\mathord{\pm}0.0006$ & $0.1311\mathord{\pm}0.0008$ & $9.690\mathord{\pm}0.054$\\
T1 & E & $0.9161\mathord{\pm}0.0004$ & $0.0979\mathord{\pm}0.0002$ & $1.323\mathord{\pm}0.085$\\
 & R & $0.9271\mathord{\pm}0.0012$ & $0.0960\mathord{\pm}0.0004$ & $0.775\mathord{\pm}0.016$\\
Apollo & E & $0.9098\mathord{\pm}0.0015$ & $0.0873\mathord{\pm}0.0001$ & $6.114\mathord{\pm}0.056$\\
 & R & $0.9552\mathord{\pm}0.0040$ & $0.0955\mathord{\pm}0.0006$ & $6.593\mathord{\pm}0.022$\\
Berkeley & E & $0.8752\mathord{\pm}0.0048$ & $0.1647\mathord{\pm}0.0001$ & $6.233\mathord{\pm}0.010$\\
 & R & $0.8906\mathord{\pm}0.0065$ & $0.1448\mathord{\pm}0.0001$ & $6.865\mathord{\pm}0.014$\\
\bottomrule\end{tabular}

\end{minipage}
\end{figure*}

\subsection{Tracing a video reference on G1}
\begin{figure*}[!t]
\centering\includegraphics[width=\textwidth]{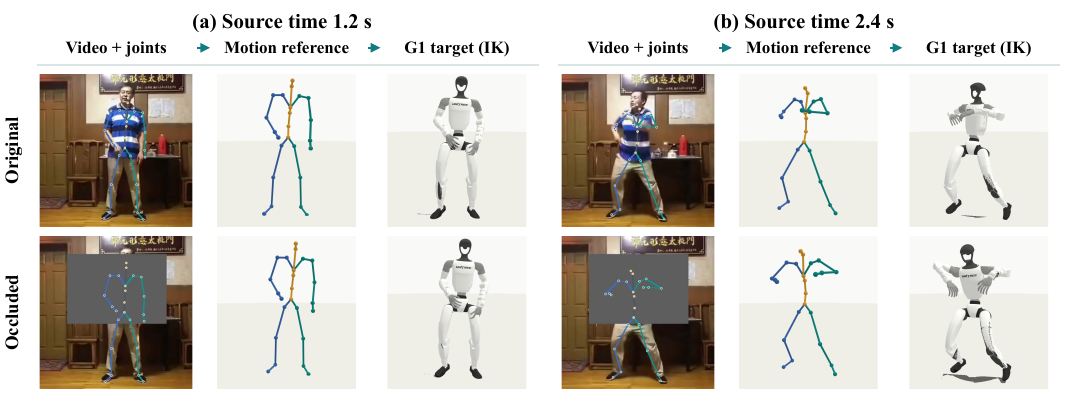}
\caption{\textbf{Video source alignment and embodiment mapping.} Original and occluded inputs at $t=1.2$ and $2.4$~s are shown with aligned motion and target joint references. Read each row from the input frame through the recovered motion to the mapped joint target for G1.}
\label{fig:video}
\par\bigskip\noindent\begin{minipage}[t]{\columnwidth}\normalfont\normalsize\raggedright At 1.2 and 2.4~s, the two rows of Fig.~\ref{fig:video} lead from the input frames to different G1 targets. Execution starts at the reference XY position and yaw. With the same pretrained tracker and its native lookahead, both runs finish the four seconds without a fall or contact outside the feet. Figure~\ref{fig:physical} follows the original input into execution. Both elbow angles drop near 2~s. Later in the clip the knees make larger excursions around their references. The curves compare the measured joints with the targets read by the controller at those source times.\end{minipage}\hfill\begin{minipage}[t]{\columnwidth}\vspace{0pt}
\captionof{table}{\tablecaptiontext{G1 physical tracking from two video input conditions. Height: MAE. XY: mean and maximum position error. One run per condition. Both conditions use the same G1 tracker.}}
\label{tab:video}\centering\small
\begin{tabular}{@{}lcccc@{}}\toprule
Input & Falls & $e_q$ (rad) & Height (cm) & XY (m)\\\midrule
Original & 0 & 0.123 & 0.98 & 0.207 / 0.678\\
Occluded & 0 & 0.133 & 1.18 & 0.352 / 1.295\\\bottomrule
\end{tabular}
\end{minipage}\par\bigskip
\centering\includegraphics[width=\textwidth]{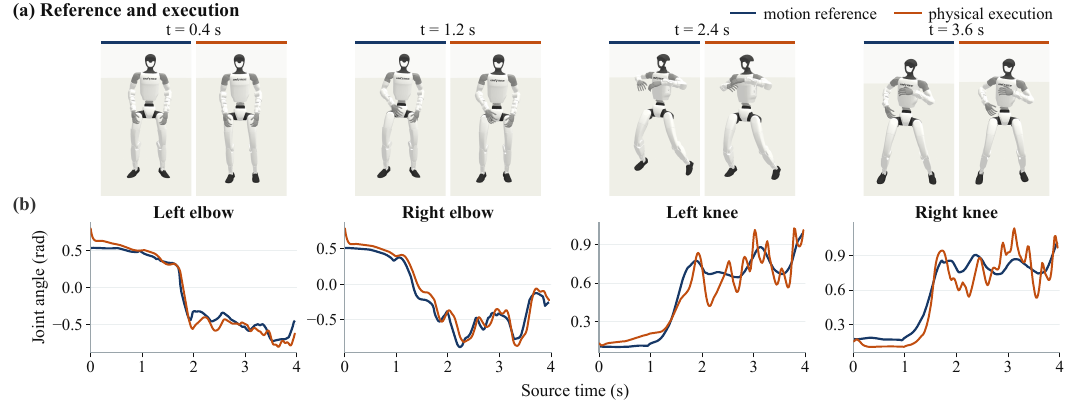}
\caption{\textbf{Dynamic execution of the recovered reference.} Joint traces use logged samples from the original input run. References retain the native sequence lookahead, and all curves preserve their source timestamps, linking the mapped target sequence to the robot's physical response. Solid navy curves mark the controller reference and solid orange curves mark physical execution.}
\label{fig:physical}
\end{figure*}

On a second clip, we pass the original and occluded references through the same pretrained G1 tracker. For the original and occluded inputs, the mean joint RMS errors relative to their respective references are $0.123$ and $0.133$~rad (Table~\ref{tab:video}). Mean horizontal position errors are $0.207$ and $0.352$~m, respectively. The maximum for the occluded input is $1.295$~m. Absolute XY position and world yaw are evaluated separately from the velocity, height, orientation, and joint channels supplied to the tracker.

\section{Discussion and Conclusion}

MotionSkill reuses the geometry and timing of a filmed step while allowing each biped to follow a joint trajectory suited to its limbs. In simulation, ten policies, each trained for one body on the same four-second walk, completed 98\% of trials from held out initial states. With partial occlusion, eight bodies received a feasible trajectory. Of their 40 trials, 13 reached the end and 27 stopped during tracking. Across four pretrained controllers, updating phase throughout a change in cadence gave higher correlation between measured and demonstrated foot separation than aligning phase at reset.


\end{document}